\pdfoutput=1

\documentclass[11pt, dvipsnames]{article}

\usepackage[preprint]{acl}

\usepackage{times}
\usepackage{latexsym}
\usepackage{algorithm} 
\usepackage{algpseudocode} 
\usepackage{amsmath}
\usepackage{xcolor}
\usepackage{tikz}
\usepackage{soul}
\usepackage{ctable}
\usepackage{amsmath}
\usepackage[most]{tcolorbox}
\usepackage{amssymb}
\usepackage{enumerate}
\usepackage{enumitem}
\usepackage{xurl} 
\usepackage[T1]{fontenc}

\usepackage[utf8]{inputenc}

\usepackage{microtype}

\usepackage{inconsolata}

\usepackage{graphicx}
\usepackage{epstopdf}

\title{What You Read Isn't What You Hear: \\Linguistic Sensitivity in Deepfake Speech Detection}

\author{
 \textbf{Binh Nguyen\textsuperscript{1}}\;\;\;\;
 \textbf{Shuju Shi\textsuperscript{2}}\;\;\;\;
 \textbf{Ryan Ofman\textsuperscript{3}}\;\;\;\;
 \textbf{Thai Le\textsuperscript{2}}
 \\
 \textsuperscript{1}Independent Researcher\;\;\;\;
 \textsuperscript{2}Indiana University\;\;\;\;
 \textsuperscript{3}Deep Media AI
\\
   \textsuperscript{1}\texttt{nqbinh17@apcs.fitus.edu.vn}\;\;\; \textsuperscript{2}\texttt{\{shi16,tle\}@iu.edu}\;\;\; \textsuperscript{3}\texttt{ryan@deepmedia.ai}
}

\begin{document}
\maketitle
\begin{abstract}
Recent advances in text-to-speech technologies have enabled realistic voice generation, fueling audio-based deepfake attacks such as fraud and impersonation. While audio anti-spoofing systems are critical for detecting such threats, prior work has predominantly focused on acoustic-level perturbations, leaving the impact of linguistic variation largely unexplored. In this paper, we investigate the linguistic sensitivity of both open-source and commercial anti-spoofing detectors by introducing transcript-level adversarial attacks. Our extensive evaluation reveals that even minor linguistic perturbations can significantly degrade detection accuracy: attack success rates surpass 60\% on several open-source detector–voice pairs, and notably one commercial detection accuracy drops from 100\% on synthetic audio to just 32\%. Through a comprehensive feature attribution analysis, we identify that both linguistic complexity and model-level audio embedding similarity contribute strongly to detector vulnerability. We further demonstrate the real-world risk via a case study replicating the Brad Pitt audio deepfake scam, using transcript adversarial attacks to completely bypass commercial detectors. These results highlight the need to move beyond purely acoustic defenses and account for linguistic variation in the design of robust anti-spoofing systems. All source code will be publicly available.

\end{abstract}

\section{Introduction}

Recent advances in text-to-speech (TTS) technology have enabled natural speech synthesis in over 7,000 languages~\cite{lux2024metalearningtexttospeechsynthesis} and high-fidelity audio from just a single sample of a target voice~\cite{chen2024f5ttsfairytalerfakesfluent}. However, these innovations have also made it easier for attackers to create deepfake audio for identity fraud, evident in a surge by more than 2,000\% over the past three years in deepfake fraud~\cite{daSilva2024aibradpitt} and the recent notorious case of Brad Pitt impersonation scam of over \$800K~\cite{signicat2024deepfake}.

To counter deepfake audio, audio anti-spoofing systems (AASs) have been developed to distinguish genuine--i.e., human-spoken speech, from spoofed one--i.e., machine-synthesized speech for identify falsification~\cite{jung2021aasistaudioantispoofingusing, tak2021endtoendantispoofingrawnet2, wu2024cladrobustaudiodeepfake, tak2022automaticspeakerverificationspoofing}. However, AAS are known to be vulnerable to \textit{acoustic-level manipulations} such as injection of small noise, volume modification, and even deliberate attacks or often so-called \textit{adversarial manipulations}~\cite{wu2024cladrobustaudiodeepfake, wu2020defenseadversarialattacksspoofing, muller2023complexvaluedneuralnetworksvoice}. Such the vulnerability is also analogous to \textit{text-level manipulations} targeting deepfake text detectors where synonym replacement of only a few words or small variations in word choice can significantly alter their detection probabilities~\cite{uchendu2023attribution}. 

\begin{figure}[tb!]
  \includegraphics[width=7cm]{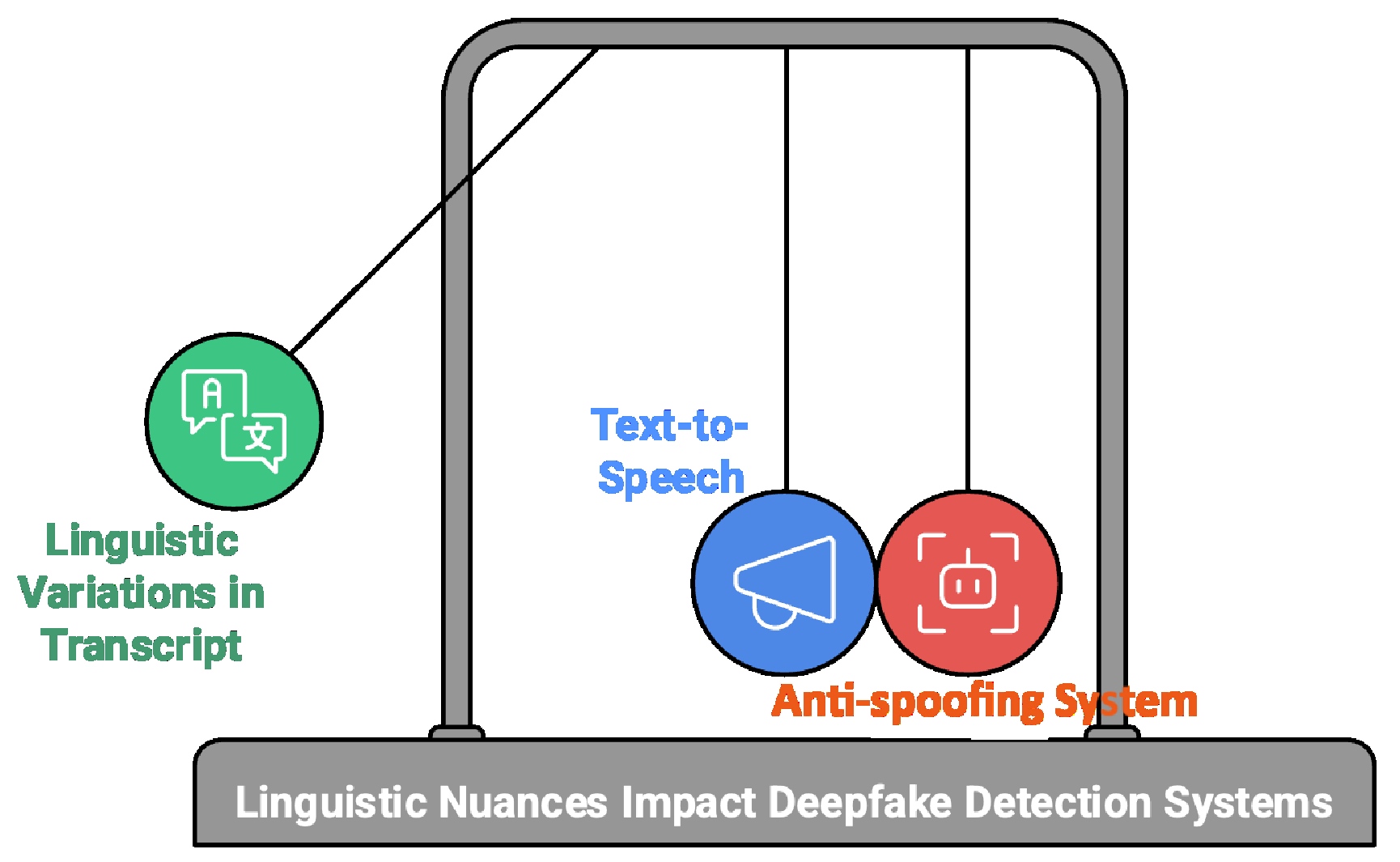}
  \caption{Linguistic variation of the transcript can swing the confidence of audio anti-spoofing system.}
  \label{fig:motivation}
  \vspace{-10pt}
\end{figure}

Since any speech fed into an AAS is either synthesized by TTS or spoken by humans from an input transcript written in human languages, it is natural to hypothesize that AASs, although only accept audio inputs, might be also \textit{indirectly influenced by text-level, linguistic manipulations} on the audio' transcripts (Fig. \ref{fig:motivation}). However, research questions such as \textit{``whether such an effect of such linguistic variations, such as word choice or dialect, on anti-spoofing performances of AAS exist?} or \textit{``when and how much linguistic variations in transcripts influence the effectiveness of audio anti-spoofing systems?''} are under-explored. 

Such questions are also intuitive and relevant to how humans perceive information from auditory speech. Particularly, linguistic differences can influence how human listeners perceive and evaluate speech, sometimes resulting in bias or negative responses~\cite{Peters17082024}. Such factors may also potentially introduce vulnerabilities or biases into AASs, especially from adversarial machine learning perspectives when they are often trained on human-curated data. For instance, terms like ``illegal'' versus ``undocumented'', although semantic neighbors, can affect a speaker's perceived credibility~\cite{Lim2018UnderstandingCO}. If such linguistic sensitivity indeed exists in current AASs, they can be manipulated by malicious actors to carefully craft spoof audio that is much more challenging for AASs to accurately detect as fake.

To examine the linguistic sensitivity of AASs, this work takes an initial step toward evaluating the central hypothesis: \textit{subtle linguistic variations within a transcript can propagate through a text-to-speech (TTS) pipeline and significantly impact the predictions of AASs} (Fig.~\ref{fig:motivation}). To this end, we formulate our investigation as an adversarial attack scenario, wherein a malicious actor strategically introduces minimal perturbations to an audio transcript, while preserving its original meaning, prior to its conversion to audio via a TTS pipeline, with the goal of evading detection by state-of-the-art (SOTA) AASs. Our empirical validation demonstrates that both research and production-grade detectors are significantly vulnerable to such subtle linguistic manipulations, with attack success rates exceeding 60\% across both open-source and commercial AASs. Moreover, we show that linguistic nuances correspond to translated acoustic qualities in the spoofed audio, ultimately affecting AASs' accuracy. \textbf{Our contributions} can be summarized as follows.




\begin{enumerate}[leftmargin=\dimexpr\parindent-0.01\labelwidth\relax,noitemsep,topsep=0pt]
\item To the best of our knowledge, our work is the first that formulates and examines linguistic sensitivity in automatic audio anti-spoofing systems.
\item We develop a transcript-level adversarial attack pipeline that generates semantically valid perturbations and demonstrates how subtle linguistic changes can degrade detection accuracy, in many cases from over 90\% to just below 20\%, in both open-source and commercial detectors.
\item We perform feature attribution analysis of over 14 linguistic, acoustic, and model-level features and analyze how they correlate with such linguistic vulnerability, offering insights for more robust audio anti-spoofing systems.
\end{enumerate}

\section{Motivation}
\subsection{Related Work}

Most adversarial attack work in audio anti-spoofing focuses on signal or acoustic-level attacks, such as noise injection or frequency masking, to expose vulnerabilities in spoof detection models~\cite{attorresi2022combiningautomaticspeakerverification, 10.1145/3543507.3583222}. However, little attention has been paid to the role of linguistic variation. As a result, efforts to improve robustness have mainly addressed acoustic distortions and cross-dataset challenges via domain adaptation and knowledge distillation~\cite{arora2022impactchannelvariationoneclass}. In contrast, specific impacts of transcript manipulations, such as what types of text perturbations and what are their effectiveness, remain underexplored in audio anti-spoofing. Whereas text-based adversarial attacks in NLP have demonstrated that small semantic changes can fool classifiers~\cite{jin2020bertreallyrobuststrong, le-etal-2022-perturbations}, it is still unknown how such linguistic perturbations, when indirectly propagated through TTS synthesis (Fig. \ref{fig:motivation}), would affect downstream audio spoofing detection.

\subsection{Preliminary Analysis}

We first carry out a preliminary analysis is to examine whether linguistic variations in transcripts might affect the robustness of AASs. To do this, we randomly substitute one word in each of 1439 transcripts with a synonym, synthesized audio for both versions with a TTS model, and test them with the high-performing open-source audio spoofing AASIST-2 detector. Surprisingly, even minimal, one-word changes cause AASIST-2 detector to misclassify up to 5.7\% of samples, and bona-fide detection probabilities drop by as much as 67.9\% in some cases (Table ~\ref{tab:prelim_examples}). Moreover, most open-source AAS are trained on the ASVSpoof-2019 LA dataset, which displays significant linguistic disparities between spoofed and bona-fide samples (Table~\ref{tab:asv_2019_stats}). Statistical tests confirm that spoofed transcripts are statistically more complex in terms of token perplexity and readability than bona-fide cases (Table~\ref{tab:t_test_asv}). Such disparities in training data can introduce linguistic bias into the trained anti-spoofing models. Motivated by these observations, we will then systematically evaluate and quantify what degree AASs are sensitive to small changes in audio' transcripts through a comprehensive, algorithmic approach.

\begin{table}[tb!]
\footnotesize
\centering
\begin{tabular}{p{5.2cm}p{1.7cm}}
\toprule
\textbf{Transcript} & \textbf{Bona-fide \%} \\
\cmidrule(lr){1-2}
She is a \textcolor{red}{\st{successful}}\textcolor{blue}{\bf good} actor with $\ldots$ & 3.6 $\xrightarrow{}$ 80.5\\
The trust was unable to \textcolor{red}{\st{pay}}\textcolor{blue}{\bf reward} the$\ldots$ & 3.7 $\xrightarrow{}$ 34.9\\
$\ldots$ for a \textcolor{red}{\st{man}}\textcolor{blue}{\bf guy} or woman of letters. & 3.7 $\xrightarrow{}$ 50.2\\
\bottomrule
\end{tabular}
\caption{A few examples of preliminary transcript-level adversarial attacks on anti-spoofing detector AASIST-2.}
\label{tab:prelim_examples}
\vspace{-10pt}
\end{table}

\section{Problem Formulation}\label{sec:formulation}
Let $\mathcal{F}{:}X{\rightarrow}Y$ be an AAS, which maps the audio input $X$ to the bona-fide label output $Y$. Given a set of $N$ transcripts $\mathcal{T}${=}$\{T_1, T_2, \ldots, T_N\}$ with $T_i{\in}\mathbb{R}^{M}$, and a TTS model $\mathcal{G}{:}T{\rightarrow}X$, we synthesize a collection of $N$ audio $\mathcal{X}${=}$\{X_1,X_2,\ldots,X_N\}$ with $X_i{\in} \mathbb{R}^{L}$, by entering each transcript in $\mathcal{T}$ to $\mathcal{G}(\cdot)$. Moreover, $\mathcal{Y}{=}\{Y_1,Y_2,\ldots,Y_N\}$ is the ground truth of $\mathcal{X}$, where $Y_i{\to}0$ indicates a spoofing label, and $Y_i{\to}1$ means a bona-fide label. In our setting, $Y_i$ is a spoofing label--i.e., $Y_i{\leftarrow}0\;\forall i$, since $X_i$ is a machine-synthesized audio. $M$ is the number of words in a transcript, and $L$ is the wavelength of an audio. 

We define an AAS $\mathcal{F}(\cdot)$ as \textbf{linguistically sensitive to a specific TTS model $\mathcal{G}(\cdot)$} if its prediction of audio synthesized via $\mathcal{G}(\cdot)$ is flipped \textit{solely by making small changes to its underlying transcript while preserving its original semantic meaning}, without modifying the audio synthesis system, the speaker profile, the acoustic characteristics directly. Formally, given a transcript $T$, $\mathcal{F}(\cdot)$ exhibits linguistic sensitivity to $\mathcal{G}(\cdot)$ if there exists a \textit{perturbed} transcript $\tilde{T}$ such that:
\begin{equation}
\begin{split}
\text{SIM}(T, \tilde{T}){==}1,\; \mathcal{F}(\mathcal{G}(\tilde{T})){==}\tilde{Y}, \text{and}\ \tilde{Y}{\neq}Y,
\end{split}
\end{equation}
\noindent where the boolean indicator $\text{SIM}(T, \tilde{T})$ ensures the adversarial transcript $\tilde{T}$ remains faithful to the original meaning or purpose (e..g, intend to transfer money) and also the original structure or syntax of $T$, such that the changes are subtle enough that a human cannot easily detect an intent to fool the system. 

\begin{table}[tb!]
  \centering  \footnotesize
  \begin{tabular}{lccc}
    \toprule
    \textbf{Metric} & \textbf{$\Delta$}  & \textbf{$t$}    & \textbf{$p(\text{one{-}sided})$} \\
    \cmidrule(lr){1-4}
    Tokens               &  1.59     & 30.70  & 0.0000 \\
    Phonemes             &  6.26     & 26.86  & 0.0000 \\
    Readability          &  0.17     & 2.12   & 0.0172 \\
    Token PPL           & 27.43     & 6.46   & 0.0000 \\
    Phoneme PPL   &  0.00     & 0.17   & 0.4345 \\
    \bottomrule
  \end{tabular}
  \caption{Results of independent two-sample t-tests comparing spoof and bona-fide items on ASVSpoof 2019 training data statistics. PPL is the perplexity.}
  \label{tab:t_test_asv}
\end{table}

To systematically find such perturbed $\tilde{T}$ or to find out if $\mathcal{F}(\cdot)$ is linguistically sensitive to $\mathcal{G}(\cdot)$, therefore, we formulate this task as an optimization problem with an objective function as follows.

\begin{tcolorbox}[
colback=white,
title=Objective Function,
fontupper=\normalsize
]
Given a transcript $T{=}\{w_1, w_2, \ldots, w_M\}$, a target AAS $\mathcal{F}$, and a TTS model $\mathcal{G}$, our goal is to find an alternative transcript $\tilde{T}$ by \textit{minimally perturbing} $T$, or:
\begin{align*}
&\;\;\;\;\tilde{T}^* = \arg\min_{\tilde{T}} \text{distance}(T, \tilde{T}^*)\;\text{s.t.}\\
&\text{SIM}(T, \tilde{T}){==}1,\; \mathcal{F}(\mathcal{G}(\tilde{T})){==}\tilde{Y}, \text{and}\ \tilde{Y}{\neq}Y
\end{align*}
\vspace{-1.5em}
\label{def:objective_function}
\end{tcolorbox}

\section{Method}

To solve the introduced optimization problem, we adapt the adversarial attack framework in adversarial NLP literature to design an adversarial transcript perturbation framework that exploits the linguistic sensitivity of the AASs. 
The overall algorithm is formalized in Alg. ~\ref{algo:transcript_generation} (Appendix). 

\vspace{2pt}
\noindent \textbf{Step 1: Finding Important Words.} First, we want to measure how sensitive $\mathcal{F}(\cdot)$'s prediction is to each word, allowing us to prioritize perturbations to the most influential locations in the transcript. To do this, given a transcript $T{=}\{ w_1, w_2, \ldots, w_m \}$, we estimate the impact of each word $w_i$ on the anti-spoofing prediction (Lines 3-5). For each word position $i$, we synthesize audio without $w_i$ in the transcript and then compute the bona-fide probability $p_i{=}\mathcal{F}(\mathcal{G}(T_{\backslash w_i}))$. Then, we aggregate all masked-transcript candidates $\mathcal{W}{=}\{T_{\backslash w_1}, ..., T_{\backslash w_m}\}$ and sort them by the descending impact scores $p_i$ (Line 6, Alg. \ref{algo:transcript_generation}).

\vspace{2pt}
\noindent \textbf{Step 2: Greedy Word Perturbations.} Next, our algorithm iteratively attempts to find effective word substitutions beginning with the positions with the largest impact, potentially minimizing the number of words needed to perturb. For each candidate position, we use a $\textit{Search}(\cdot)$ to propose a set of replacement candidates (Line 9). To search for the replacement $\tilde{w}_i$ for $w_i$, $\textit{Search}(\cdot)$ first finds a list of replacement candidates by utilizing either WordNet to find synonym replacements~\cite{ren-etal-2019-generating}, or a Masked Language Model ~\cite{devlin2019bertpretrainingdeepbidirectional} that masks the word $w_i$ as the token $[MASK]$ and performs the token predictions ~\cite{jin2020bertreallyrobuststrong}. 
For each substitution candidate $c_k$, we then construct a new transcript $T'$ by replacing $w_i$ in $\tilde{T}$ with $c_k$ (Line 10). The new transcript is then synthesized into speech using $\mathcal{G}(T')$ and evaluated by the anti-spoofing model to obtain an updated bona-find prediction score $p'$ (Line 11). The best replacement candidate $c_k$ is one that best maximizes $p'$ while satisfying all constraints listed in Eq. (\ref{def:objective_function}) (Line 12-15, Alg. \ref{algo:transcript_generation}).

\vspace{2pt}
\noindent \textbf{Step 3: Semantic and Syntax Preservation.} We factorize the boolean check $\text{SIM}(T, \tilde{T})$ (Eq. \ref{def:objective_function}) to (1) syntactic and (2) semantic preservation. To check for syntactic preservation, we only accept a replacement $\tilde{w}_i$ only if its part-of-speech (POS) function in $\tilde{T}$ preserves that of $w_i$ in the original transcript $T$. To check for semantic preservation, we ensure that the cosine similarity, denoted as $\text{cos}(\cdot)$, between semantic vectorized representations of $T$ and the current $\tilde{T}$ embed using the popular Universal Sentence Encoder (USE) ~\cite{cer2018universalsentenceencoder}, denoted as $f_\text{embed}(\cdot)$, is at least $\delta$ threshold:
\begin{equation}
\text{cos}\left(f_\text{embed}(T), f_\text{embed}(\tilde{T})\right) \geq \delta
\label{def:semantic}
\end{equation}



\noindent \textbf{Overall}, our framework is \textit{model-agnostic} and does \textit{not} require access to internal parameters or gradients of the target AAS, making it applicable in practical black-box settings where we only have query access to the AAS. By leveraging linguistic variability at the transcript level while enforcing functional constraints, the proposed method is able to systematically probe and exploit the linguistic sensitivity of end-to-end audio anti-spoofing detectors.

\section{Experiments}
\subsection{Set-up}

\noindent \textbf{Datasets.} We evaluate our algorithm on a total of 1,439 test transcripts from the deepfake speech VoiceWukong dataset~\cite{yan2024voicewukongbenchmarkingdeepfakevoice}, statistics of which is provided in Table~\ref{tab:voice_wukong_stats}. VoiceWukong is constructed based on the English VCTK dataset~\cite{yamagishi2019vctk} and is released under the Creative Commons Attribution-NonCommercial 4.0 International Public License, with which we comply by using the data exclusively for research purposes. To better reflect real-world attack scenarios, we restrict our evaluation to transcripts having at least 10 tokens.

\vspace{2pt}
\noindent \textbf{Anti-spoofing Detectors.} We use three open-source anti-spoofing models: AASIST-2 ~\cite{tak2022automaticspeakerverificationspoofing}, CLAD~\cite{wu2024cladrobustaudiodeepfake}, RawNet-2~\cite{tak2021endtoendantispoofingrawnet2}, and two commercially available deepfake speech detection APIs, which will be refereed using pseudonyms API-A and API-B.

Table~\ref{tab:detector_acc_voice_wukong} presents the precision and recall scores for bona-fide and spoof prediction across all models. Notably, AASIST-2 achieves the most balanced and robust detection, with high precision and recalls. In contrast, CLAD and RawNet-2 show comparatively lower and more variable performance. Commercial detectors exhibit much lower bona-fide recall, indicating a tendency to misclassify legitimate speech as spoofed.

\renewcommand{\tabcolsep}{2pt}
\begin{table*}[tb!]
    \centering
    \footnotesize
    \begin{tabular}{lcccccccccccc}
        \toprule
         \textbf{Text-to-Speech} & \multicolumn{4}{c}{\textbf{AASIST-2}} & \multicolumn{4}{c}{\textbf{CLAD}} & \multicolumn{4}{c}{\textbf{RawNet-2}} \\
         & \textbf{OC}$\uparrow$ & \textbf{AUA}$\downarrow$ & \textbf{ASR}$\uparrow$ & \textbf{COS}$\uparrow$ & \textbf{OC}$\uparrow$ & \textbf{AUA}$\downarrow$ & \textbf{ASR}$\uparrow$ & \textbf{COS}$\uparrow$ & \textbf{OC}$\uparrow$ & \textbf{AUA}$\downarrow$ & \textbf{ASR}$\uparrow$ & \textbf{COS}$\uparrow$\\
         \cmidrule(lr){2-5} \cmidrule(lr){6-9} \cmidrule(lr){10-13}
        Kokoro (British Male)  & 95.1\% & 39.9\% & \textbf{58.1\%} & 90.5\%  & 96.9\% & 43.9\% & 54.7\% & 91.7\%  & 100.0\% & 98.9\% & 1.1\% & 89.5\% \\
Kokoro (British Female)  & 92.4\% & 30.4\% & \textbf{67.1\%} & 91.3\%  & 97.7\% & 62.4\% & 36.2\% & 90.4\%  & 92.4\% & 56.9\% & 38.4\% & 90.4\% \\
Kokoro (American Male)  & 90.4\% & 37.9\% & \textbf{58.0\%} & 91.1\%  & 83.3\% & 43.3\% & 56.6\% & 91.5\%  & 100.0\% & 100.0\% & 0.0\% & 88.2\% \\
Kokoro (American Female)  & 92.2\% & 32.6\% & 64.6\% & 91.0\%  & 98.3\% & 32.6\% & 66.9\% & 92.1\%  & 83.2\% & 20.3\% & \textbf{75.6\%} & 93.3\% \\
\cmidrule(lr){1-13}
Coqui (Donald Trump)  & 85.5\% & 25.7\% & \textbf{69.9\%} & 93.2\%  & 98.9\% & 62.0\% & 37.3\% & 92.3\%  & 99.8\% & 88.4\% & 11.4\% & 90.2\% \\
Coqui (Elon Musk)  & 98.0\% & 75.9\% & 22.5\% & 90.9\%  & 98.4\% & 62.1\% & \textbf{36.9\%} & 91.1\%  & 100.0\% & 99.9\% & 0.1\% & 89.7\% \\
Coqui (Taylor Swift)  & 94.4\% & 17.0\% & \textbf{82.0\%} & 92.9\%  & 95.1\% & 20.0\% & 79.0\% & 92.8\%  & 99.7\% & 88.2\% & 11.5\% & 89.5\% \\
Coqui (Oprah Winfrey)  & 98.9\% & 79.0\% & \textbf{20.1\%} & 91.5\%  & 99.7\% & 99.7\% & 0.0\% & 90.5\%  & 95.8\% & 86.2\% & 9.9\% & 91.0\% \\
\cmidrule(lr){1-13}
F5 (Male)  & 88.5\% & 33.4\% & 62.2\% & 92.8\%  & 93.2\% & 7.9\% & \textbf{91.6\%} & 94.6\%  & 99.4\% & 78.0\% & 21.5\% & 90.7\% \\
        \bottomrule
    \end{tabular}
    \caption{Open-source model results. Bold values indicate the TTS-voice pair that is most effective at attacking (ASR) each detector model.}
    \label{table:opensource}
    \vspace{-10pt}
\end{table*}

\vspace{2pt}
\noindent \textbf{Text-to-Speech Models.} We employ Kokoro TTS, a lightweight high-quality, and community-known model, capable of generating $10K$ audio in only $832$ seconds on an NVIDIA A100 GPU. For voice cloning TTS, we use Coqui TTS (having over 39K stars on GitHub) to replicate the voices of four well-known individuals. Additionally, we evaluate F5 TTS ~\cite{chen2024f5ttsfairytalerfakesfluent}, a recently proposed SOTA model with best-in-class generation quality. For commercial TTS, we employ OpenAI TTS due to its popularity and low cost.

\vspace{2pt}
\noindent \textbf{Word Perturbation Methods.} We adapt four word perturbation strategies for Step 2 of Alg. \ref{algo:transcript_generation} and TextAttack framework~\cite{morris2020textattack}, including PWWS~\cite{ren-etal-2019-generating}, which swaps words using WordNet; TextFooler~\cite{jin2020bertreallyrobuststrong}, which substitutes words based on contextual word embeddings while respecting part-of-speech and filtering stop words; BAE~\cite{garg-ramakrishnan-2020-bae}, which leverages BERT to propose plausible replacements; and BERTAttack~\cite{li-etal-2020-bert-attack}, which also uses BERT to generate adversarial substitutions. These strategies cover most of the word perturbation methods in adversarial NLP literature.

\vspace{2pt}
\noindent \textbf{Metrics.} We report the Original Accuracy (OC), Accuracy Under Attack (AUA) of the target AAS on the synthesized spoof samples. We also measure Attack Success Rate (ASR) or the percentage of spoof audio out of the tested transcripts that were able to flip the original correct spoof predictions of the target AAS detector. We also report the semantic preservation score (COS) calculated via Eq. (\ref{def:semantic}) and standardized to 0-100\% scale. Intuitively, the higher the ASR, the lower AUA, and the higher the COS, the better an attack is able to preserve the original transcripts' meaning.

\textit{We refer the readers to the Appendix for additional implementation details.}

\subsection{Results}


\subsubsection{Attacking Open-Source Detectors}
Table~\ref{table:opensource} summarizes the average performance of three open-source anti-spoofing detectors: AASIST-2, CLAD, and RawNet-2 under adversarial attacks on synthetic speech generated from a variety of TTS models across four word perturbation strategies (PWWS, TextFooler, BAE, and BERTAttack), totaling 108 experiments.

\vspace{2pt}
\noindent \textbf{Overall Linguistic Sensitivity.} All three detectors show a marked reduction in AUA, suggesting that adversarially perturbed transcripts can noticeably degrade anti-spoofing performance while consistently maintaining high semantic preservation close to or higher than 90\%. Consequently, ASR is often substantial, reaching as high as 82\%, specially for certain voice profiles. 

\vspace{2pt}
\noindent \textbf{Voice Gender Effect.} Overall, female voices exhibit a higher ASR than male voices across both detectors and TTS systems. For example, for Kokoro TTS voices, British Female and American Female identities consistently yield higher ASRs than their male counterparts, often accompanied by sharply lower AUA. This implies that spoof female voices are more prone to become undetected under linguistic adversarial manipulations.

\vspace{2pt}
\noindent \textbf{Notable Exceptions.} Some voice profiles are notably resistant to attack. For instance, Coqui TTS voice for Oprah Winfrey shows almost zero ASR on both CLAD (0.02\%), but this phenomenon does not repeat with other detectors. Similarly, the RawNet-2 detector demonstrates strong robustness to some male voice profiles, such as Kokoro TTS (British Male and American Male) and Coqui Elon Musk voice cloning, where the ASR only reaches (1.06\% and 0.00\%) and 0.14\%, respectively, indicating that linguistic sensitivity of an AAS is TTS-specific and \textit{some detector-voice combinations are far less susceptible to transcript-based attacks}. This also validates our AAS-TTS pair linguistic sensitivity formulation in Sec. \ref{sec:formulation}. We later show that these voice-detector combinations have nearly perfect Audio Encoder Similarity (Fig. ~\ref{fig:feature_image}), meaning that audio encoders of the TTS and the detector are more or less encoding similar information.

\subsubsection{Attacking Commercial Detectors}

Table~\ref{table:commercial} presents the attack results on commercial AASs when paired with both commercial and non-commercial TTS models. To conserve API usage and cost, each experiment applies only the strongest attack method identified in prior experiments (TextFooler), and evaluates 100 items that were randomly sampled to maintain the same length distribution as the main test set. For each TTS-detector pair, we attack the voice profile with the highest original accuracy (OC) to demonstrate the lower bound effectiveness in the hardest-case scenario. For OpenAI's TTS, we choose CLAD which has the highest original accuracy among the open-source models.

\renewcommand{\tabcolsep}{5pt}
\begin{table}[tb!]
    \footnotesize
    \centering
    \begin{tabular}{lcccc}
        \toprule
        & \textbf{OC}$\uparrow$ & \textbf{AUA}$\downarrow$ & \textbf{ASR}$\uparrow$ & \textbf{COS}$\uparrow$ \\
        \cmidrule(lr){2-5}
        API-A - Coqui & 100.0\% & 98.0\% & 2.0\% & 85.7\% \\
API-A - F5 & 99.0\% & 70.0\% & 29.3\% & 86.2\% \\
API-A - Kokoro & 100.0\% & 74.0\% & 26.0\% & 84.1\% \\
API-A - OpenAI & 95.0\% & 32.0\% & 66.3\% & 89.3\% \\
\cmidrule(lr){1-5}
API-B - Coqui & 100.0\% & 100.0\% & 0.0\% & 87.0\% \\
API-B - F5 & 100.0\% & 100.0\% & 0.0\% & 87.3\% \\
API-B - Kokoro & 100.0\% & 100.0\% & 0.0\% & 80.8\% \\
API-B - OpenAI & 100.0\% & 96.0\% & 4.0\% & 86.3\% \\
\cmidrule(lr){1-5}
CLAD - OpenAI & 86.0\% & 4.0\% & 95.3\% & 93.4\% \\
        \bottomrule
    \end{tabular}
    \caption{Commercial Anti-spoofing Detectors Results}
    \label{table:commercial}
    \vspace{-10pt}
\end{table}

For API-A, we observe a substantial drop in detection accuracy under attack when pairing with most TTS models except Coqui. While its OC is nearly perfect across all voices, adversarial attack reduces AUA to as low as 32\% when paired with OpenAI TTS, resulting in a high attack success rate (ASR) of 66.3\%. Notably, API-A is vulnerable when tested with realistic, high-quality TTS synthesis. API-B, in contrast, retains perfect detection (AUA = 100\%) for Coqui, F5, and Kokoro TTS, and only exhibits a minor decrease (AUA = 96\%, ASR = 4\%) with OpenAI TTS. However, Table~\ref{tab:detector_acc_voice_wukong} reveals this robustness is partly due to a strong bias toward labeling all samples as spoof, with poor bona-fide recall and moderate spoof precision. For the open-source CLAD model evaluated on OpenAI TTS, adversarial attack drops the accuracy from 86\% to just 4\%, yielding the highest ASR (95.3\%) among all tested scenarios. These findings highlight the concerning vulnerability to linguistic sensitivity of commercial detectors faced when with high-fidelity synthetic speech.

\section{Feature Analysis}

Beyond providing empirical validation on our initial hypothesis that audio anti-spoof detectors are sensitive to small linguistic changes in the audio' underlying transcripts, this section aims to investigate and analyze \textit{what factors and how much they associate with varying anti-spoofing detectors' decisions under adversarial attacks}. Particularly, we extract \textbf{linguistic, acoustic, and model-level features} from 108 open-source attack experiments, utilize logistic regression analysis, and train predictive models to estimate the bona-fide probability of perturbed inputs. Formulations of all features are provided in the Appendix.

\subsection{Feature Engineering}
We first seek to understand how \textbf{linguistic features (LF)} at transcript-level shift under adversarial attacks. 

\vspace{2pt}
\noindent \textbf{LF1. Perturbed Percentage} measures the fraction of modified words in a transcript; higher values indicate more extensive lexical changes.

\vspace{2pt}
\noindent \textbf{LF2. Readability Difference} quantifies the change in reading comprehension difficulty between the original and perturbed transcripts using the Dale-Chall Readability Score.

\vspace{2pt}
\noindent \textbf{LF3. Semantic Similarity} assesses the similarity in meaning between the original and perturbed transcripts using Universal Sentence Encoder embeddings, or the COS evaluation metric. 

\vspace{2pt}
\noindent \textbf{LF4. Perplexity Difference} measures the change in perplexity between the original and perturbed transcripts.

\vspace{2pt}
\noindent \textbf{LF5. Syntactic Complexity Difference} measures the change in maximum syntactic tree depth between the original and perturbed transcripts.

Given that text-level changes can propagate to measurable differences at the acoustic level, we further investigate how variations in several \textbf{acoustic features (AF)} contribute to the performance of anti-spoofing detectors.

\vspace{2pt}
\noindent \textbf{AF1. DTW Distance} utilizes Dynamic Time Warping to measure the alignment cost between the mel spectrograms of the original and perturbed audio. 

\vspace{2pt}
\noindent \textbf{AF2. Duration Difference} captures the difference in audio length.

\vspace{2pt}
\noindent \textbf{AF3. Phoneme Perplexity Difference} measures the corresponding change in phoneme sequence perplexity, calculated via the CharsiuG2P~\cite{zhu2022byt5modelmassivelymultilingual} T5-based model.

\vspace{2pt}
\noindent \textbf{AF4. Aesthetics Difference} measures the shifting aesthetics computed by Meta Audiobox Aesthetics ~\cite{tjandra2025metaaudioboxaestheticsunified} which includes four automatic quality assessment measures: Content Enjoyment (CE), Content Usefulness (CU), Production Complexity (PC), Production Quality (PQ).

\renewcommand{\tabcolsep}{2pt}
\begin{table}[tb!]
    \centering
    \footnotesize
    \begin{tabular}{lcccc}
        \toprule
         & \textbf{coef} & \textbf{std err}& \textbf{z} & \textbf{P>|z|}\\
         \cmidrule(lr){1-5}
Perturbed Percentage & -0.3507 & 0.009 & -39.31 & 0.000 \\
$\Delta$ Readability  & 0.0352 & 0.007 & 4.69 & 0.000 \\
$\Delta$ PPL & 0.0758 & 0.009 & 8.48 & 0.000 \\
$\Delta$ Tree Depth & 0.0125 & 0.007 & 1.76 & 0.077 \\
\cmidrule(lr){1-5}
$\Delta$ Duration  & 0.0748 & 0.008 & 8.89 & 0.000 \\
DTW Distance & 0.2063 & 0.008 & 26.73 & 0.000 \\
$\Delta$ Phoneme PPL & -0.0112 & 0.007 & -1.54 & 0.122 \\
$\Delta$ Content Enjoyment & 0.0536 & 0.013 & 4.18 & 0.000 \\
$\Delta$ Content Usefulness & 0.0679 & 0.021 & 3.17 & 0.001 \\
$\Delta$ Production Complexity & 0.0110 & 0.010 & 1.12 & 0.264 \\
$\Delta$ Production Quality & 0.0307 & 0.017 & 1.82 & 0.069 \\
\cmidrule(lr){1-5}
Audio Encoder Similarity & -0.9013 & 0.011 & -81.60 & 0.000 \\
Spoof F1 & 3.1254 & 0.159 & 19.60 & 0.000 \\
Bona-fide F1 & -2.5911 & 0.159 & -16.25 & 0.000 \\
        \bottomrule
    \end{tabular}
    \caption{Logistic regression feature analysis for bona-fide detection on adversarial samples. $\Delta$ is the difference and Semantic Similarity feature is removed due to the high Variance Inflation Factor to avoid multicollinearity.}
\label{table:logist_regression}
    \label{table:logist_regression}
    \vspace{-10pt}
\end{table}

\begin{figure*}[tb!]
    \centering
    \setlength{\fboxsep}{0pt}
    \includegraphics[width=16cm]{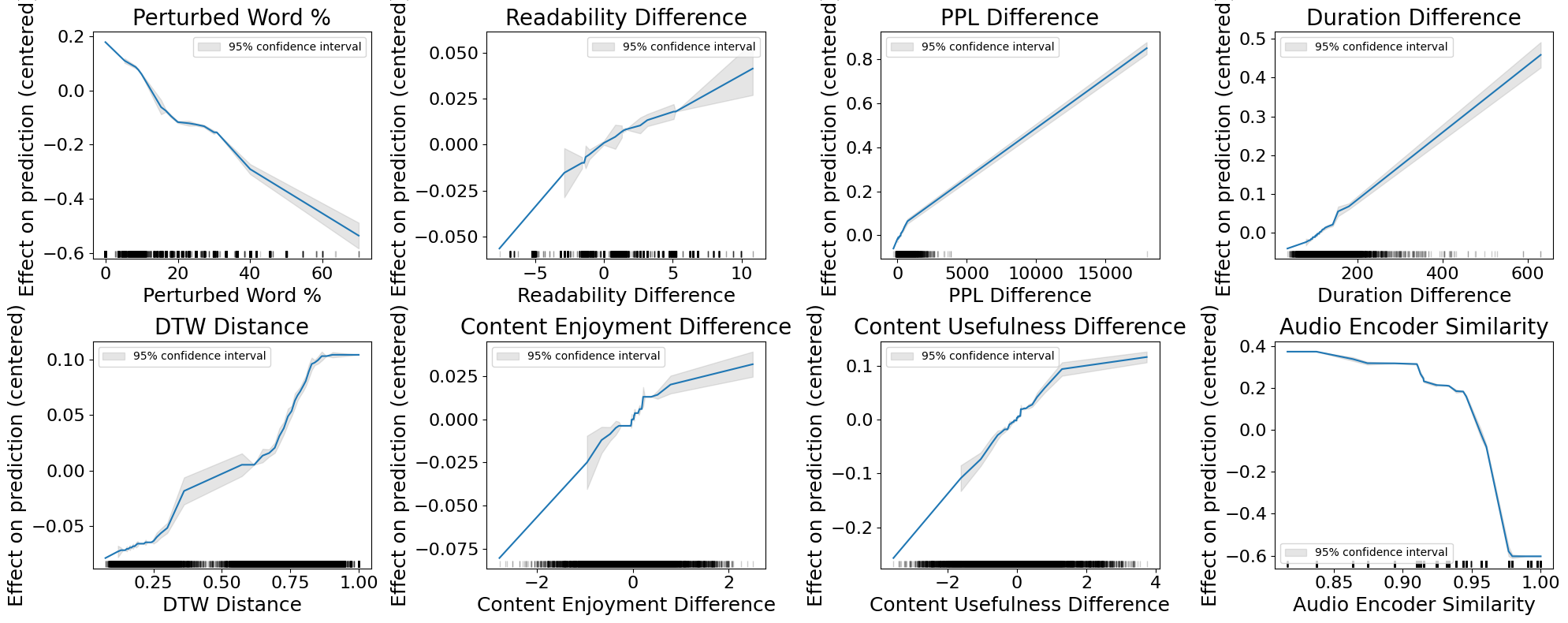}
    \caption{Feature impact on bona-fide probability prediction. A positive effect means the feature increases the likelihood of a perturbed item being classified as bona-fide.}
    \label{fig:feature_image}
\end{figure*}

For \textbf{model-level features (MF)}, we propose:

\vspace{2pt}
\noindent \textbf{MF1. Audio Encoder Similarity(AES)} metric quantifies how closely synthesized audios of the same voice cluster in the detector's representation space. A high AES score indicates that the detector perceives all TTS-generated audio for a given voice as acoustically similar, or being able to capture them as originally from the same voice profile, which may enhance robustness against transcript-level adversarial attacks. 

\vspace{2pt}
\noindent \textbf{MF2, MF3. Spoof and bona-fide F1.} Additionally, we include spoofed and bona-fide F1 scores (Table \ref{tab:detector_acc_voice_wukong}) as model-level features to analyze how pre-existing biases influence behavior under adversarial attacks. Notably, if these features can predict attack outcomes, they are especially useful because they can be computed \textbf{before} any adversarial perturbation, guiding the selection or development of more robust anti-spoofing models.

\subsection{Analysis Results}

Table~\ref{table:logist_regression} presents the results of a logistic regression analysis predicting the bona-fide probability for adversarial audio samples using the engineered features.

\vspace{2pt}
\noindent \textbf{Linguistic Features Impact.} Several features display significant associations with the detector’s response to adversarial perturbations with statistical significance. Notably, the proportion of perturbed words in a transcript is negatively correlated with bona-fide detection, indicating that increasing lexical modifications decreases the likelihood that the detector classifies the input as bona-fide. Syntactic complexity differences are less significant, indicating that deep syntactic restructurings are less impactful than surface-level wording and fluency changes. PPL difference and readability difference are both positively correlated with bona-fide probability. The trends in Fig.~\ref{fig:feature_image} suggests that when the perturbed transcript exhibits greater linguistic complexity than the original, the adversarial sample is more likely to be classified as bona-fide. This leads to an assumption that the disparities in linguistic features between spoofed and bona-fide training samples (Table~\ref{tab:asv_2019_stats}) might have introduced linguistic vulnerabilities that can be exploited by adversarial attack algorithms.

\begin{table*}[tb!]
\scriptsize
\footnotesize
\centering
\begin{tabular}{lp{13cm}l}
\toprule
 & \textbf{Transcript} & \textbf{Bona-fide} \\
\cmidrule(lr){1-3}
Fraud: & Anne, I need to \textcolor{red}{\st{be}}\textcolor{blue}{\bf become} direct. $\ldots$ I need your help \textcolor{red}{\st{immediately}}\textcolor{blue}{\bf succinctly}. & 0.2 $\xrightarrow{}$ \textbf{69.7}\\
Victim: & Brad? What is it? You sound serious.& N/A\\
Fraud: & I'm in the \textcolor{red}{\st{hospital}}\textcolor{blue}{\bf convalescent}. It's serious. \textcolor{red}{\st{Kidney}}\textcolor{blue}{\bf Liver} cancer. They \textcolor{red}{\st{need}}\textcolor{blue}{\bf crucial} to $\ldots$ & 0.2 $\xrightarrow{}$ \textbf{62.4}\\
Victim: & Cancer? Oh no, Brad, I'm so sorry to hear that. What kind of problem with funds? Don't you?& N/A\\
Fraud: & My accounts are frozen. $\ldots$ the \textcolor{red}{\st{courts}}\textcolor{blue}{\bf judiciary} have tied up everything$\ldots$ \textcolor{red}{\st{hospital}}\textcolor{blue}{\bf outpatient} bill.& 0.4 $\xrightarrow{}$ \textbf{90.3}\\
Fraud: & The doctors need \textcolor{red}{\st{payment}}\textcolor{blue}{\bf reimbursement} now to proceed with this vital step. $\ldots$& 0.3 $\xrightarrow{}$ \textbf{73.1}\\
Victim: & Me? But I... I'm not a millionaire, Brad. How much do they need?& N/A\\
Fraud: & $\ldots$ It's 830,000 euros. I \textcolor{red}{\st{know}}\textcolor{blue}{\bf understand}
 it's a huge ask, Anne, but my life could depend on this.& 0.7 $\xrightarrow{}$ \textbf{58.2}\\
Victim: & 830,000 euros?! That's an enormous sum. But how would I even? And where would I send it?& N/A\\
\bottomrule
\end{tabular}
\caption{Illustration of how adversarial transcript attacks on Brad Pitt voice cloning scam enable the attacker to significantly undermine a commercial audio anti-spoof detector.}
\label{tab:case_study clone}
\end{table*}

\vspace{2pt}
\noindent \textbf{Acoustic Features Impact.} The DTW distance between mel-spectrograms and the duration difference indicate that greater spectral or temporal deviations between original and perturbed audio samples are associated with higher bona-fide probabilities (trends in Fig.~\ref{fig:feature_image}). In contrast, the effect of phoneme perplexity difference is not statistically significant, suggesting that changes in phoneme-level predictability are less associated with variations in acoustic realization, such as spectral and durational differences. The positive correlations observed for Content Enjoyment and Content Usefulness suggest that enhanced emotional and artistic qualities in perturbed audio may increase its likelihood of deceiving anti-spoofing detectors.

\vspace{2pt}
\noindent \textbf{Model-Level Features Impact.} AES provides the strongest predictive signals for susceptibility to adversarial attacks. AES is negatively associated with bona-fide prediction, implying that models that produce highly clustered audio embeddings for a given TTS and voice are less likely to recognize perturbed inputs as bona-fide. Notably, as shown in Fig.~\ref{fig:feature_image}, when AES approaches 100\%, there is a significant reduction in the likelihood of attack success. Additionally, the original detector F1 scores on bona-fide and spoofed samples have the highest-magnitude coefficients, indicating that initial model bias in spoof discrimination translates directly into vulnerability or robustness under adversarial conditions. 

\renewcommand{\tabcolsep}{4pt}
\begin{table}[tb!]
\footnotesize
\centering
\begin{tabular}{lccc}
\toprule
\textbf{Model} & \textbf{Precision} & \textbf{Recall} & \textbf{F1 Score} \\
\cmidrule(lr){1-4}
Logistic Regression    & 66.82\%& 71.54\% & 69.10\% \\
XGBoost                & 74.70\% & 78.39\% & 76.50\% \\
Random Forest          & 64.28\% & 76.88\% & 70.02\% \\
SVM (poly kernel)    & 46.83\% & 97.00\% & 63.16\% \\
LightGBM & 72.98\% & 77.00\% & 74.94\%\\
AdaBoost & 70.30\% & 77.82\% & 73.87\%\\
\bottomrule
\end{tabular}
\caption{Performance of approximating anti-spoof detector's bona-fide prediction of various predictive models built on our engineered features.}
\label{tab:pred_models}
\vspace{-15pt}
\end{table}

Table \ref{tab:pred_models} further shows how predictive models built on our engineered features can effectively approximate the outcomes of the detector's decisions. Notably, XGBoost achieves an F1 score of up to 76.50\%, with several other models performing in the 69\% to 74\% range. This shows that these models can enable the development of grey-box or black-box adversarial attacks where attacker access to the actual detector is restricted or limited. By optimizing transcript modifications based on proxy predictions from the predictive models, adversaries can effectively attack audio anti-spoofing systems even without full transparency of the detector, underscoring the urgent need for more robust and resilient defenses.

\section{Case Study: Deepfake Voice Cloning}

Table~\ref{tab:case_study clone} presents a simulated case study adapting from the recent, notorious Brad Pitt impersonation scam where adversarial perturbations are applied to the fraudster's dialogue generated by ChatGPT and synthesized using the SOTA F5 TTS voice-cloned model. The transcripts illustrate the original and perturbed words, with corresponding bona-fide detection probabilities reported from API-A commercial anti-spoofing detector. In all tested exchanges, the unperturbed fraud utterances are assigned extremely low bona-fide probabilities ($<1\%$), suggesting effective spoof detection. However, after targeted adversarial perturbation of key lexical items (e.g., ``be''${\rightarrow}$``become'', ``hospital''${\rightarrow}$``convalescent''), the bona-fide probability rises dramatically, with post-attack scores ranging from 58.2\% to as high as 90.3\%. Notably, even minimal lexical changes can evade commercial detectors, flipping the label from clear spoof to likely bona-fide.

These findings demonstrate the concerning real-world risks of transcript-level adversarial attacks in voice cloning scenarios, highlighting the urgency for developing more robust anti-spoofing mechanisms that can withstand subtle semantic and lexical manipulations.

\section{Conclusion}
This work demonstrate that SOTA audio anti-spoofing systems are vulnerable to transcript-level linguistic nuances. By systematically applying semantic preserving perturbations to transcripts, we show that even subtle linguistic changes can significantly degrade detection accuracy in both open-source and commercial deepfake detectors. Our experiments and feature analyses reveal that these vulnerabilities are driven by both linguistic complexity and characteristics of the model’s learned audio representations. This underscores the need for anti-spoofing systems to consider linguistic variation, not just acoustics. For future work, we plan to further investigate the interplay between model architecture, training data, and linguistic features that contribute to adversarial susceptibility, with the goal of guiding more comprehensive and resilient detection strategies.

\newpage
\clearpage
\section*{Limitation}
One limitation of our work is the lack of experimentation on false positive cases, such as those involving non-native speakers who may stutter or use incorrect wording during conversations. These effects can act as natural adversarial attacks on the transcript and potentially reduce bona-fide detection accuracy. Additionally, vocalizations such as laughter, giggling, and chuckling, which may enhance the enjoyment and naturalness of generated audio, are not addressed in this study; these elements could also serve as another modality for transcript-based adversarial attacks.

Our experiments are primarily limited to English-language data, leaving open the question of how linguistic attacks generalize to other languages and multilingual TTS systems. Diverse syntactic and morphological structures in non-English languages may uniquely impact anti-spoofing system robustness, which remains unexplored in this work.

Furthermore, although the linguistic perturbations in our experiments are constrained to retain semantic meaning, we do not measure their detectability by humans or plausibility in real conversational contexts. User studies are needed to assess whether such adversarial modified transcripts sound unnatural or prompt suspicion among human listeners.

\section*{Broader Impacts and Potential Risks}
By uncovering vulnerabilities related to linguistic perturbations, our findings encourage audio anti-spoofing research to move beyond acoustic analysis and incorporate linguistic robustness into system design and evaluation. This insight can directly inform the development of safer, more resilient voice authentication and verification products.

Our methodology highlights the importance of adversarial testing and “red teaming” in the responsible development of AI security systems. This proactive approach enables the community to identify and mitigate attacks before they are exploited in practice, ultimately safeguarding critical voice-driven infrastructure.

This research is conducted to advance audio security and raise awareness of vulnerabilities in current anti-spoofing systems. The authors are committed to promoting social good and responsible AI development, with no intention of enabling any malicious or unethical applications of these findings.

However, by publicly revealing specific attack strategies and demonstrating their effectiveness, our work could inadvertently lower the barrier for malicious actors to evade anti-spoofing systems. Additionally, making our adversarial techniques and code openly available—while essential for reproducibility and further research—introduces the risk that these methods might be misused for fraudulent purposes.

\newpage
\clearpage
\bibliography{custom}

\newpage
\clearpage
\appendix

\setcounter{table}{0}
\renewcommand{\thetable}{A\arabic{table}}

\section{Dataset statistics}
\subsection{ASVSpoof 2019 statistics}
Table~\ref{tab:asv_2019_stats} summarizes the key linguistic and structural statistics of the ASVSpoof 2019 LA training dataset, segmented by speaker gender (male and female) and ground-truth label (spoof vs. bona-fide). For each group, we report the average number of tokens and phonemes per utterance, the average readability score (which reflects the linguistic complexity of the transcripts), and perplexity values computed at both the token and phoneme levels. Notably, all reported values are statistically significant ($p$-value~$<$~0.001), suggesting consistent differences between spoof and bona-fide samples across these linguistic features. These statistics provide critical insight into the dataset composition, which may influence both the performance and the generalization capacity of anti-spoofing models during training and evaluation.

\begin{table}[tb]
    \centering
    \scriptsize
    \footnotesize
    \begin{tabular}{lcccc}
    \toprule
        & \multicolumn{2}{c}{\textbf{Male Voice}} & \multicolumn{2}{c}{\textbf{Female Voice}}\\
         & \textbf{Spoof} & \textbf{Bona-fide} & \textbf{Spoof} & \textbf{Bona-fide}\\
        \cmidrule(lr){2-3}\cmidrule(lr){4-5}
    $\mathbb{E}[Tokens]$      & 7.85   & 7.08   & 9.09  & 6.92  \\
    $\mathbb{E}[Phonemes]$    & 30.09  & 27.26  & 35.56 & 26.89 \\
    $\mathbb{E}[Readability]$ & 6.27   & 6.26   & 6.89  & 6.60  \\
    $\mathbb{E}[TokenPPL]$   & 100.12 & 96.18  & 94.58 & 96.20 \\
    $\mathbb{E}[PhonePPL]$ & 1.0461 & 1.0458 & 1.0460 & 1.0460\\
    \bottomrule
    \end{tabular}
    \caption{ASVSpoof 2019 LA statistics. All values is statistically significant ($p-value$ < 0.001). $\mathbb{E}[\cdot]$ is the average value of that metric, and PPL is perplexity.}
    \label{tab:asv_2019_stats}
\end{table}

\subsection{VoiceWukong dataset statistics}
Table~\ref{tab:voice_wukong_stats} presents key statistical features of the VoiceWukong dataset used in our experiments. In addition to average transcript length in tokens and phonemes, we report average readability scores, which provide an estimate of the linguistic complexity of the dataset's transcripts, as well as token-level and phoneme-level perplexity (PPL), which serve as measures of sequence unpredictability. These features offer a comprehensive overview of the structural and linguistic properties of the evaluation data, and help contextualize the challenges involved for both TTS synthesis and anti-spoofing detection.

\begin{table}[tb]
    \centering
    \footnotesize
    \begin{tabular}{lc}
    \toprule
         \textbf{Feature} & \textbf{Value} \\
         \cmidrule(lr){1-2}
         Average Tokens & 11.05\\
         Average Phonemes & 42.46\\
         Average Readability & 7.28\\
         Token PPL & 43.86\\
         Phoneme PPL & 1.0497\\
    \bottomrule
    \end{tabular}
    \caption{VoiceWukong dataset features such as readability and perplexity.}
    \label{tab:voice_wukong_stats}
\end{table}

\section{Equations and Results}
\label{sec:appendix}

\subsection{Implementation Details}
For open-source detectors, instead of fine-tuning models for each specific TTS-generated voice, we adapt them using batch normalization calibration~\cite{shomron2020posttrainingbatchnormrecalibration} on a small set that does not overlap the evaluation data, which shifts the mean and variance of the feature distributions to better match those of the current TTS system until detection accuracy exceeds 90\%. We argue that retraining on every possible voice is infeasible, given the potentially over 8 billion unique speakers worldwide; however, these voices are likely to share similar statistical properties in their acoustic features.

\subsection{Anti-spoof detection performance on VoiceWukong dataset}
Table~\ref{tab:detector_acc_voice_wukong} compares the performance of various anti-spoofing detectors on the VoiceWukong dataset, reporting both precision and recall for bona-fide and spoofed audio. We evaluate two commercial APIs (API-A and API-B) alongside several state-of-the-art open-source models: RawNet-2, CLAD, and AASIST-2. The results reveal considerable variation in performance across different systems. Notably, AASIST-2 achieves the highest and most balanced precision and recall scores for both bona-fide and spoofed classes, suggesting superior generalization capability. In contrast, the commercial detectors—especially API-B—exhibit strong bias, with high spoof recall but low bona-fide recall, indicating a tendency to label most samples as spoofed. These findings highlight the strengths and limitations of existing anti-spoofing solutions on challenging synthetic datasets, and motivate the need for further robustness against linguistic and generative variation.

\begin{table}[tb]
    \centering
    \footnotesize
    \begin{tabular}{lcccc}
    \toprule
        & \multicolumn{2}{c}{\textbf{Bona-fide}} & \multicolumn{2}{c}{\textbf{Spoof}} \\
        \textbf{Detector} & \textbf{Precision} & \textbf{Recall} & \textbf{Precision} & \textbf{Recall} \\
        \cmidrule(lr){2-3}\cmidrule(lr){4-5}
        
         API-A &  80.9\% & 63.0\% & 58.2\% & 77.5\% \\
         API-B &  90.0\% & 28.1\% & 46.8\% & 95.3\% \\
         \cmidrule(lr){1-5}

         RawNet-2 & 89.1\% & 63.0\% & 61.3\% & 88.3\% \\
         CLAD & 88.4\% & 66.7\% & 63.4\% & 86.8\% \\
         AASIST-2 & 90.3\% & 90.9\% & 86.1\% & 85.4\% \\
    \bottomrule
    \end{tabular}
    \caption{Anti-spoof detection performance comparison.}
    \label{tab:detector_acc_voice_wukong}
    \vspace{-10pt}
\end{table}

\subsection{Transcript-level Adversarial Attack Algorithm}
Our transcript-level adversarial attack Algorithm~\ref{algo:transcript_generation} identifies the most influential words in a target transcript and greedily substitutes them—using synonym replacement or masked language models—with alternatives that maximize the chance of misclassification by the anti-spoofing system, while ensuring both semantic and syntactic fidelity through embedding similarity and part-of-speech checks. This model-agnostic, black-box framework exploits the linguistic sensitivity of audio anti-spoofing systems without requiring access to internal model parameters, demonstrating the practical risks posed by transcript-level adversarial perturbations.

\begin{algorithm}[tb!]
    \caption{Adversarial Transcript Generation}
    \label{algo:transcript_generation}
    \begin{algorithmic}[1]
        \State \textbf{Input:} A transcript $T = \{ w_1, w_2, \ldots, w_m \}$, the audio anti-spoofing detection  $\mathcal{F}(\cdot)$, a Text-to-Speech model $\mathcal{G}(\cdot)$, $SeachMethod$ and $Constraints$

        \State \textbf{Output:} Adversarial transcript $\tilde{T}$

        \State Identify the impact $p_i$ of a word $w_i$
        
        \For {$w_i$ in $w_1, w_2, \ldots, w_m$}
            \State $p_i$ $\gets$ $\mathcal{F}(\mathcal{G}(T_{\backslash w_i}))$
        \EndFor
        
        \State $\mathcal{W}$ $\gets$ $\{T_{\backslash w_1}, T_{\backslash w_2}, \ldots, T_{\backslash w_m}\}$, sorted by descending values of $p_i$

        \State $\tilde{T}$ $\gets$ $T$,\;\;$\tilde{p}$ $\gets$ $\mathcal{F}(\mathcal{G}(\tilde{T}))$
        \For {$T_{\backslash w_i}$ in $\mathcal{W}$}
            \State $\text{Candidates}$ $\gets$ $\text{Search}(T_{\backslash w_i})$
            \For{$c_k$ in ${\text{Candidates}}$}
                \State $T'$ $\gets$ Replace $c_k$ with $w_i$ in $\tilde{T}$
                \State $p'$ $\gets$ $\mathcal{F}(\mathcal{G}(T'))$
                \If{SIM($T$, $\tilde{T}$) \textbf{AND} $p'$ > $\tilde{p}$}
                  \State $\tilde{T}$ $\gets$ $T'$,\;\;$\tilde{p}$ $\gets$ $p'$
                \EndIf 
            \EndFor
        \EndFor
    \State \textbf{return} $\tilde{T}$
    \end{algorithmic} 
\end{algorithm}

\subsection{Additional Results}
We provide experimental results for AASIST-2 in Table ~\ref{appendix:aasist2}, CLAD in Table ~\ref{appendix:clad}, and Rawnet-2 in Table ~\ref{appendix:rawnet2}.

\begin{table}[tb!]
\footnotesize
\begin{tabular}{cccccc}
  \hline
    \textbf{Voice} & \textbf{Method} & \textbf{OC} & \textbf{AUA} & \textbf{ASR} & \textbf{COS} \\
  \cmidrule(lr){2-6}
Donald Trump & BAE & 85.5 & 27.3 & 68.0 & 93.3 \\
Donald Trump & BertAttack & 85.5 & 21.6 & 74.7 & 93.7 \\
Donald Trump & PWWS & 85.5 & 35.7 & 58.2 & 94.4 \\
Donald Trump & TextFooler & 85.5 & 18.3 & 78.6 & 91.3 \\
\hline
Elon Musk & BAE & 98.0 & 78.3 & 20.1 & 91.7 \\
Elon Musk & BertAttack & 98.0 & 76.3 & 21.4 & 92.3 \\
Elon Musk & PWWS & 98.0 & 79.8 & 19.1 & 93.6 \\
Elon Musk & TextFooler & 98.0 & 70.3 & 28.3 & 87.5 \\
\hline
Oprah Winfrey & BAE & 98.9 & 79.0 & 20.1 & 91.8 \\
Oprah Winfrey & BertAttack & 98.9 & 77.1 & 22.4 & 92.5 \\
Oprah Winfrey & PWWS & 98.9 & 86.2 & 12.9 & 94.1 \\
Oprah Winfrey & TextFooler & 98.9 & 71.9 & 27.3 & 88.6 \\
\hline
Taylor Swift & BAE & 94.4 & 21.5 & 77.2 & 92.7 \\
Taylor Swift & BertAttack & 94.4 & 6.2 & 93.4 & 94.2 \\
Taylor Swift & PWWS & 94.4 & 32.9 & 65.2 & 93.8 \\
Taylor Swift & TextFooler & 94.4 & 7.5 & 92.0 & 91.0 \\
\hline
F5 \textcolor{NavyBlue}{Male} & BAE & 88.5 & 32.9 & 62.8 & 92.8 \\
F5 \textcolor{NavyBlue}{Male} & BertAttack & 88.5 & 27.4 & 69.0 & 93.6 \\
F5 \textcolor{NavyBlue}{Male} & PWWS & 88.5 & 41.5 & 53.1 & 94.4 \\
F5 \textcolor{NavyBlue}{Male} & TextFooler & 88.5 & 31.9 & 64.0 & 90.3 \\
\hline
American \textcolor{Lavender}{Female} & BAE & 92.2 & 37.3 & 59.5 & 91.7 \\
American \textcolor{Lavender}{Female} & BertAttack & 92.2 & 26.9 & 70.8 & 92.3 \\
American \textcolor{Lavender}{Female} & PWWS & 92.2 & 52.5 & 43.1 & 92.6 \\
American \textcolor{Lavender}{Female} & TextFooler & 92.2 & 13.9 & 84.9 & 87.4 \\
\hline
American \textcolor{NavyBlue}{Male} & BAE & 90.4 & 40.3 & 55.4 & 92.3 \\
American \textcolor{NavyBlue}{Male} & BertAttack & 90.4 & 35.7 & 60.5 & 92.0 \\
American \textcolor{NavyBlue}{Male} & PWWS & 90.4 & 58.6 & 35.2 & 93.1 \\
American \textcolor{NavyBlue}{Male} & TextFooler & 90.4 & 17.2 & 81.0 & 87.0 \\
\hline
British \textcolor{Lavender}{Female} & BAE & 92.4 & 39.2 & 57.6 & 92.2 \\
British \textcolor{Lavender}{Female} & BertAttack & 92.4 & 22.4 & 75.7 & 92.1 \\
British \textcolor{Lavender}{Female} & PWWS & 92.4 & 47.5 & 48.5 & 93.2 \\
British \textcolor{Lavender}{Female} & TextFooler & 92.4 & 12.5 & 86.5 & 87.8 \\
\hline
British \textcolor{NavyBlue}{Male} & BAE & 95.1 & 41.8 & 56.0 & 91.5 \\
British \textcolor{NavyBlue}{Male} & BertAttack & 95.1 & 33.8 & 64.4 & 91.4 \\
British \textcolor{NavyBlue}{Male} & PWWS & 95.1 & 62.1 & 34.8 & 92.3 \\
British \textcolor{NavyBlue}{Male} & TextFooler & 95.1 & 21.9 & 77.0 & 86.9 \\

  \hline
\end{tabular}
\caption{Complete experimental results on AASIST-2 detector}
\label{appendix:aasist2}
\end{table}

\begin{table}[tb!]
\footnotesize
\begin{tabular}{ cccccc}
  \hline
    \textbf{Voice} & \textbf{Method} & \textbf{OC} & \textbf{AUA} & \textbf{ASR} & \textbf{COS} \\
 \cmidrule(lr){2-6}

Donald Trump & BAE & 98.8 & 63.7 & 35.6 & 91.7 \\
Donald Trump & BertAttack & 98.8 & 48.2 & 51.3 & 92.1 \\
Donald Trump & PWWS & 99.0 & 74.2 & 25.1 & 93.0 \\
Donald Trump & TextFooler & 99.0 & 46.2 & 53.6 & 91.7 \\
\hline
Elon Musk & BAE & 98.5 & 69.8 & 29.1 & 91.9 \\
Elon Musk & BertAttack & 98.1 & 47.5 & 51.6 & 91.4 \\
Elon Musk & PWWS & 98.5 & 80.4 & 18.4 & 93.7 \\
Elon Musk & TextFooler & 98.5 & 50.7 & 48.5 & 87.5 \\
\hline
Oprah Winfrey & BAE & 99.7 & 99.7 & 0.0 & 91.3 \\
Oprah Winfrey & BertAttack & 99.7 & 99.7 & 0.0 & 91.0 \\
Oprah Winfrey & PWWS & 99.7 & 99.7 & 0.1 & 93.3 \\
Oprah Winfrey & TextFooler & 99.7 & 99.7 & 0.0 & 86.4 \\
\hline
Taylor Swift & BAE & 95.1 & 24.3 & 74.4 & 92.9 \\
Taylor Swift & BertAttack & 95.1 & 14.4 & 84.8 & 93.6 \\
Taylor Swift & PWWS & 95.1 & 34.5 & 63.7 & 93.8 \\
Taylor Swift & TextFooler & 95.1 & 6.6 & 93.1 & 91.1 \\
\hline
F5 \textcolor{NavyBlue}{Male} & BAE & 93.2 & 9.8 & 89.5 & 94.5 \\
F5 \textcolor{NavyBlue}{Male} & BertAttack & 93.2 & 2.9 & 96.9 & 95.4 \\
F5 \textcolor{NavyBlue}{Male} & PWWS & 93.2 & 16.6 & 82.2 & 94.5 \\
F5 \textcolor{NavyBlue}{Male} & TextFooler & 93.2 & 2.1 & 97.7 & 93.9 \\
\hline
American \textcolor{Lavender}{Female} & BAE & 98.3 & 37.9 & 61.4 & 92.3 \\
American \textcolor{Lavender}{Female} & BertAttack & 98.3 & 27.3 & 72.2 & 92.8 \\
American \textcolor{Lavender}{Female} & PWWS & 98.3 & 48.0 & 51.2 & 93.8 \\
American \textcolor{Lavender}{Female} & TextFooler & 98.3 & 17.1 & 82.6 & 89.7 \\
\hline
American \textcolor{NavyBlue}{Male} & BAE & 99.8 & 64.5 & 35.4 & 91.1 \\
American \textcolor{NavyBlue}{Male} & BertAttack & 99.8 & 58.9 & 41.0 & 90.9 \\
American \textcolor{NavyBlue}{Male} & PWWS & 34.0 & 0.0 & 100.0 & 98.8 \\
American \textcolor{NavyBlue}{Male} & TextFooler & 99.8 & 49.8 & 50.1 & 85.2 \\
\hline
British \textcolor{Lavender}{Female} & BAE & 97.7 & 67.3 & 31.1 & 91.1 \\
British \textcolor{Lavender}{Female} & BertAttack & 97.7 & 57.8 & 40.8 & 91.0 \\
British \textcolor{Lavender}{Female} & PWWS & 97.7 & 77.7 & 20.5 & 93.4 \\
British \textcolor{Lavender}{Female} & TextFooler & 97.7 & 46.6 & 52.3 & 86.0 \\
\hline
British \textcolor{NavyBlue}{Male} & BAE & 96.9 & 48.4 & 50.0 & 92.2 \\
British \textcolor{NavyBlue}{Male} & BertAttack & 96.9 & 39.2 & 59.5 & 91.8 \\
British \textcolor{NavyBlue}{Male} & PWWS & 96.9 & 59.1 & 39.0 & 93.6 \\
British \textcolor{NavyBlue}{Male} & TextFooler & 96.9 & 28.9 & 70.2 & 89.2 \\

\hline
\end{tabular}
\caption{Complete experimental results on CLAD detector}
\label{appendix:clad}
\end{table}

\begin{table}[tb!]
\footnotesize
\begin{tabular}{cccccc}
  \hline
    \textbf{Voice} & \textbf{Method} & \textbf{OC} & \textbf{AUA} & \textbf{ASR} & \textbf{COS} \\
 \cmidrule(lr){2-6}
Donald Trump & BAE & 99.8 & 88.6 & 11.2 & 91.2 \\
Donald Trump & BertAttack & 99.8 & 87.5 & 12.7 & 89.8 \\
Donald Trump & PWWS & 99.8 & 91.2 & 8.6 & 93.0 \\
Donald Trump & TextFooler & 99.8 & 85.3 & 14.5 & 86.3 \\
\hline
Elon Musk & BAE & 100.0 & 99.9 & 0.1 & 90.8 \\
Elon Musk & BertAttack & 100.0 & 99.9 & 0.1 & 91.5 \\
Elon Musk & PWWS & 100.0 & 99.9 & 0.1 & 93.1 \\
Elon Musk & TextFooler & 100.0 & 99.8 & 0.2 & 85.2 \\
\hline
Oprah Winfrey & BAE & 95.8 & 86.7 & 9.4 & 91.9 \\
Oprah Winfrey & BertAttack & 95.8 & 84.1 & 12.2 & 91.4 \\
Oprah Winfrey & PWWS & 95.8 & 90.8 & 5.2 & 93.7 \\
Oprah Winfrey & TextFooler & 95.8 & 83.4 & 12.9 & 87.0 \\
\hline
Taylor Swift & BAE & 99.7 & 93.3 & 7.4 & 92.3 \\
Taylor Swift & BertAttack & 99.7 & 82.1 & 18.3 & 87.5 \\
Taylor Swift & PWWS & 99.7 & 94.8 & 4.9 & 93.2 \\
Taylor Swift & TextFooler & 99.7 & 81.7 & 18.1 & 85.8 \\
\hline
F5 \textcolor{NavyBlue}{Male} & BAE & 99.4 & 79.1 & 20.5 & 91.3 \\
F5 \textcolor{NavyBlue}{Male} & BertAttack & 99.4 & 69.3 & 30.3 & 91.3 \\
F5 \textcolor{NavyBlue}{Male} & PWWS & 99.4 & 86.0 & 13.6 & 93.6 \\
F5 \textcolor{NavyBlue}{Male} & TextFooler & 99.4 & 77.7 & 21.9 & 86.3 \\
\hline
American \textcolor{Lavender}{Female} & BAE & 83.2 & 21.9 & 73.7 & 93.9 \\
American \textcolor{Lavender}{Female} & BertAttack & 83.2 & 16.1 & 80.6 & 94.1 \\
American \textcolor{Lavender}{Female} & PWWS & 83.2 & 36.1 & 56.7 & 94.4 \\
American \textcolor{Lavender}{Female} & TextFooler & 83.2 & 7.0 & 91.6 & 90.7 \\
\hline
American \textcolor{NavyBlue}{Male} & BAE & 100.0 & 100.0 & 0.0 & 88.6 \\
American \textcolor{NavyBlue}{Male} & BertAttack & 100.0 & 100.0 & 0.0 & 90.6 \\
American \textcolor{NavyBlue}{Male} & PWWS & 100.0 & 100.0 & 0.0 & 92.9 \\
American \textcolor{NavyBlue}{Male} & TextFooler & 100.0 & 100.0 & 0.0 & 83.5 \\
\hline
British \textcolor{Lavender}{Female} & BAE & 92.4 & 62.2 & 32.7 & 91.8 \\
British \textcolor{Lavender}{Female} & BertAttack & 92.4 & 45.4 & 50.9 & 91.0 \\
British \textcolor{Lavender}{Female} & PWWS & 92.4 & 75.0 & 18.7 & 93.2 \\
British \textcolor{Lavender}{Female} & TextFooler & 92.4 & 45.1 & 51.2 & 85.7 \\
\hline
British \textcolor{NavyBlue}{Male} & BAE & 100.0 & 99.4 & 0.6 & 90.6 \\
British \textcolor{NavyBlue}{Male} & BertAttack & 100.0 & 96.8 & 3.2 & 90.1 \\
British \textcolor{NavyBlue}{Male} & PWWS & 100.0 & 99.9 & 0.1 & 92.9 \\
British \textcolor{NavyBlue}{Male} & TextFooler & 100.0 & 99.6 & 0.4 & 84.5 \\
\hline
\end{tabular}
\caption{Complete experimental results on Rawnet-2 detector}
\label{appendix:rawnet2}
\end{table} 

\subsection{Linguistic Feature Equations}
Equation~\ref{eq:perturbed_percentage}:
$\rho_{\text{perturbed}}$ quantifies the ratio of words that have been perturbed in the transcript.

\begin{equation}
\rho_{\text{perturbed}}  = \frac{\mathrm{N_{perturbed}\ }}{\mathrm{N_{words}}}
\label{eq:perturbed_percentage}
\end{equation}

Equation~\ref{eq:readability_diff}:
$\Delta_{read}$ measures the change in transcript readability after perturbation.
\begin{equation}
    \Delta_{read} = \mathrm{read}_{\mathrm{perturbed}} - \mathrm{read}_{\mathrm{original}}
    \label{eq:readability_diff}
\end{equation}

Equation~\ref{eq:semantic_similarity}:
$\mathrm{sim_{semantic}}$ computes the cosine similarity between the semantic embeddings of the perturbed and original transcripts.
\begin{equation}
\mathrm{sim_{semantic}} = \mathrm{cosine}(Emb_{\mathrm{perturbed}}, Emb_{\mathrm{original}})
\label{eq:semantic_similarity}
\end{equation}

Equation~\ref{eq:ppl_diff}:
$\Delta_{PPL}$ captures the difference in language model perplexity before and after perturbation, as measured by Llama 3.
\begin{equation}
\Delta_{PPL}= \mathrm{PPL}_{\mathrm{perturbed}} - \mathrm{PPL}_{\mathrm{original}}
\label{eq:ppl_diff}
\end{equation}

Equation~\ref{eq:depth_diff}:
$\Delta_{syntactic}$ reflects the change in syntactic tree depth between the perturbed and original transcripts.
\begin{equation}
\Delta_{syntactic}= \mathrm{Depth}_{\mathrm{perturbed}} - \mathrm{Depth}_{\mathrm{original}}
\label{eq:depth_diff}
\end{equation}

\subsection{Acoustic Feature Equations}
Equation~\ref{eq:dtw}:
$\mathrm{dtw_{distance}}$ calculates the dynamic time warping (DTW) distance between the mel-spectrograms of the perturbed and original audio.
\begin{equation}
\mathrm{dtw_{distance}} = \mathrm{DTW}(mel_{\mathrm{perturbed}}, mel_{\mathrm{original}})
\label{eq:dtw}
\end{equation}

Equation~\ref{eq:duration_diff}:
$\Delta_{duration}$ shows the change in duration between the perturbed and original synthesized speech.
\begin{equation}
\Delta_{duration}= D_{\mathrm{perturbed}} - D_{\mathrm{original}}
\label{eq:duration_diff}
\end{equation}

Equation~\ref{eq:phoneme_ppl_diff}:
$\Delta_{PhonePPL}$ measures the change in phoneme-level perplexity after transcript perturbation.
\begin{equation}
\begin{split}
\Delta_{PhonePPL} &= \mathrm{PhonePPL}_{\mathrm{perturbed}} - \\       &\mathrm{PhonePPL}_{\mathrm{original}}
\end{split}
\label{eq:phoneme_ppl_diff}
\end{equation}

Equation~\ref{eq:aesthetics_diff}:
$\Delta_{CE}$, $\Delta_{CU}$, $\Delta_{PC}$, and $\Delta_{PQ}$ represent the changes in various audio aesthetics metrics—clarity, continuity, pronunciation correctness, and prosody quality—due to the perturbation.
\begin{equation}
\begin{split}
\Delta_{CE}&= \mathrm{CE}_{\mathrm{perturbed}} -        \mathrm{CE}_{\mathrm{original}}\\
\Delta_{CU} &= \mathrm{CU}_{\mathrm{perturbed}} -        \mathrm{CU}_{\mathrm{original}}\\
\Delta_{PC} &= \mathrm{PC}_{\mathrm{perturbed}} -        \mathrm{PC}_{\mathrm{original}}\\
\Delta_{PQ} &= \mathrm{PQ}_{\mathrm{perturbed}} -        \mathrm{PQ}_{\mathrm{original}}
\end{split}
\label{eq:aesthetics_diff}
\end{equation}

\subsection{Audio Encoder Similarity Equation}
We first extract acoustic embeddings for all original transcripts by the detector. We then compute the centroid embedding by averaging these embeddings and normalizing the result to unit length:
\begin{equation}
\mathbf{c} = \frac{1}{N} \sum_{i=1}^{N} \mathbf{e}_i, \qquad
\mathbf{\hat{c}} = \frac{\mathbf{c}}{|\mathbf{c}|}
\label{eq:aes_part_1}
\end{equation}
where $\mathbf{e}_i$ denotes the embedding for the $i$-th sample and $N$ is the total number of samples in the group.

The Audio Encoder Similarity for the group is then defined as the average cosine similarity between the centroid $\mathbf{\hat{c}}$ and each sample embedding:
\begin{equation}
\mathrm{AES} = \frac{1}{N} \sum_{i=1}^{N} \cos(\mathbf{\hat{e}}_i, \mathbf{\hat{c}})
\label{eq:aes_part_2}
\end{equation}

\end{document}